\documentclass{svproc}
\usepackage{graphicx}
\usepackage{url}

\usepackage{multirow}
\usepackage{arydshln}
\usepackage{amsmath}
\usepackage{enumitem}

\begin{document}
\mainmatter              
\title{A Deep Neural Network for Multiclass Bridge Element Parsing in Inspection Image Analysis}
\titlerunning{8WCSCM}  
%
\author{Chenyu Zhang$^1$ \and Muhammad Monjurul Karim$^1$ \and Zhaozheng Yin$^2$ \and Ruwen Qin$^1$}
\authorrunning{Zhang et al.} 
%
\tocauthor{Chenyu Zhang, Muhammad Monjurul Karim, Zhaozheng Yin, and Ruwen Qin\thanks{corresponding author}}
\institute{1. Department of Civil Engineering\\
2. Department of Biomedical Informatics, Department of Computer Science, AI Institute\\Stony Brook University, Stony Brook, NY 11794, USA\\
\email{ruwen.qin@stonybrook.edu}}

\maketitle              

\begin{abstract}
Aerial robots such as drones have been leveraged to perform bridge inspections. Inspection images with both recognizable structural elements and apparent surface defects can be collected by onboard cameras to provide valuable information for the condition assessment. This article aims to determine a suitable deep neural network (DNN) for parsing multiclass bridge elements in inspection images. An extensive set of quantitative evaluations along with qualitative examples show that High-Resolution Net (HRNet) possesses the desired ability. With data augmentation and a training sample of 130 images, a pre-trained HRNet is efficiently transferred to the task of structural element parsing and has achieved a 92.67\% mean F1-score and 86.33\% mean IoU.


\end{abstract}
\section{Introduction}
%

Traditional manual bridge inspection requires a crew of inspectors, heavy equipment with lifting capacity, access to dangerous heights, and road closure during the inspection. Besides, manual bridge inspection results are subjective, varying from one inspector to another even though they follow the best inspection practice. Limitations of the traditional approach motivate research into automating bridge monitoring and evaluation with advanced technologies such as robotics and image analysis.

\begin{figure}[tb]
\centering
\includegraphics[width=12cm]{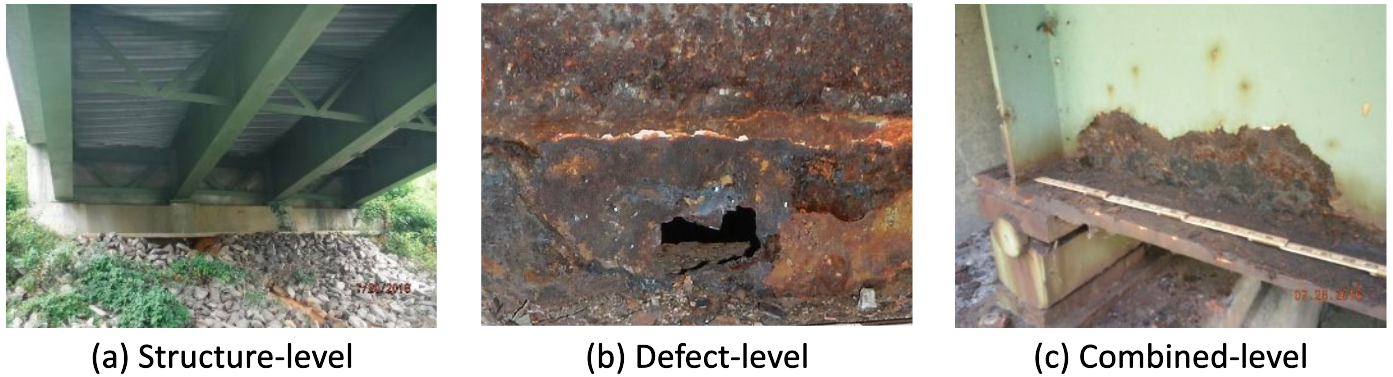}
\caption{Different levels of inspection images}
\centering
\label{Tab:DifferentLevels}
\end{figure}

Researchers have developed segmentation DNNs to detect and segment bridge elements in the inspection images because it is essential for computer vision-based bridge inspection data analysis \cite{Zhao:Chen,Karim:Yin}. According to the AASHTO’s Bridge Element Inspection Manual \cite{AASHTO}, structural elements and their defects must be associated to produce an overall rating for a whole bridge. Fig. \ref{Tab:DifferentLevels} illustrates images that contain different levels of detail about bridges. Some datasets focus on structure-level images wherein multiple structural elements are salient objects. Others are defect-level images that only contain pixel-level details of surface defects. Yet, cameras also capture bridge images from a certain distance where images have recognizable structural elements and apparent surface defects on the elements. Such inspection images provide valuable information for the condition assessment. In those images, bridge elements of the same type have widely different appearances due to the imperfect and changing view of the camera(s). Multiple types of structural elements may have similar textures and close contact. The bridge also mixes with the cluttered, dynamic background in inspection images. Therefore, multiclass bridge element parsing in inspection images or videos remains challenging. This paper aims to determine a suitable deep learning network for this purpose.

\section{Methodology}
\subsection{Design of Experiments}
This study implemented multiple latest segmentation networks and identified the HRNet \cite{hrnet} (HRNetV2-W32) as a suitable DNN for structural element parsing, supported by numerical experiments and qualitative examples. For this purpose, four experiments were conducted. The first experiment evaluated the transferability of a pre-trained HRNetV2-W32 to the task of bridge element parsing. The second experiment compared HRNetV2 of different backbones to other DNNs. The third experiment evaluated the impact of training sample size, and the last experiment examined the effects of data augmentation and class imbalance problem. The class-wise performance of HRNetV2-W32 and qualitative examples further revealed its strengths and weaknesses.


\subsection{Dataset}
The study covered 145 images, a portion of the COCO-Bridge 2021+ dataset \cite{coco}. This project used image annotation tool LabelMe \cite{label} to provide the pixel-level annotation of six common classes of structural elements for steel girder bridges: Bearing (Brg), Bracing (Brc), Deck (Dck), Floor beam (Flb), Girder (Grd), and Substructure (Sbt). In total, 822 instances are annotated in the 145 images. The study reserves 130 images for training models and 15 images for testing. Three sizes of training dataset (70, 100, and 130 images) were used to evaluate the impact of training sample size. Table \ref{Tab:SampleDistribution} further summarizes the distribution of instances by class. 
\begin{table}[htbp]
\caption{Distribution of the Structural Elements by Class}
\label{Tab:SampleDistribution}
\centering
\begin{tabular}{l|@{\quad}r|@{\quad}r@{\quad}r@{\quad}r@{\quad}r@{\quad}r@{\quad}r|@{\quad}r}
\hline\rule{-2pt}{12pt}
\multirow{2}{*}{}&\multicolumn{1}{r|@{\quad}}{No.}&\multicolumn{7}{c}{No. structural elements}\\
\cline{3-9}\rule{0pt}{10pt}
&\multicolumn{1}{r|@{\quad}}{images}&\multicolumn{1}{@{\quad}r@{\quad}}{Brg}&\multicolumn{1}{r@{\quad}}{Brc}&\multicolumn{1}{r@{\quad}}{Dck}&\multicolumn{1}{r@{\quad}}{Flb}&\multicolumn{1}{r@{\quad}}{Grd}&\multicolumn{1}{r|@{\quad}}{Sbt}&\multicolumn{1}{@{\quad}r}{Total count}\\
\hline\rule{-2pt}{10pt}
Training & 130  & 169  & 76  &  104  & 84 & 217 & 99 &749 \\
\;\!Testing  & 15  & 19  & 6  &  6  & 10 & 20 & 12 &73 \\
\hdashline[2.5pt/2pt]\rule{-2pt}{10pt}
Total count   & 145 & 188  & 82  & 110  & 94 & 237 & 111 &822\\
\hline
\multicolumn{1}{l}{\,Proportion}& &.23 & .10  & .13  & .11  & .29 & .14 & 1.00\\[1pt]
\hline
\end{tabular}
\end{table}

\subsection{Model Training Approach}
Training and testing were performed using one Nvidia Tesla V100 GPU with 32 GB of memory. Stochastic gradient descent was used as the optimizer and the momentum was 0.9. The cosine annealing scheduler was utilized with a maximum learning rate of 0.004 and a minimum of 0.00004. A batch size of 8 was used in this training process.

\subsection{Performance Metrics}
To evaluate the segmentation performance of each network, four metrics are calculated at the class level: {\it Precision} (the portion of pixels predicted to be a class that are predicted correctly), {\it Recall} (the portion of pixels in a class that are predicted correctly), {\it F1-score} (the harmonic mean of Precision and Recall), and {\it Intersection over Union} (IoU) (the number of pixels common between the ground-truth and prediction masks divided by the total number of pixels present across both masks). Then, the average values across classes were further calculated (i.e., macro-level mean values), including mean Precision (mPrecision), mean Recall (mRecall), mean F1-score (mF1), and mean IoU (mIoU).

%


%
\section{Results and Discussions}

%
\subsection{Efficiency of Transfer Learning}

Transfer learning is an effective method to address the problem of small training data, for example due to the difficulty in data collection. In the meantime, a transferable network can save training time compared to building a model from scratch. The study would like to evaluate the transferability of a pre-trained HRNetV2 to the task of structural element parsing, as well as the efficiency of transfer learning. Three training strategies were compared: training a HRNetV2-W32 from the scratch, refining the entire network of the HRNetV2-W32 pre-trained on the PASCAL Visual Object Classes (VOC) 2012 dataset and Semantic Boundaries Dataset (SBD), and refining only the rear-end portion of HRNetV2-W32 by freezing the backbone. The training sample size for this experiment is 70 images. The comparison summarized in Table \ref{Tab:TransferLearning} demonstrates the transferability of the pre-trained HRNetV2 to the task of multiclass bridge structural element parsing and so its efficiency.


\begin{table}[htbp]
\caption{Cost-effectiveness of Transfer Learning}
\centering
\begin{tabular}{l@{\quad}|r@{\quad}r@{\quad}r@{\quad}r@{\quad}rl}
\hline\rule{-2pt}{12pt}
\multirow{2}{*}{Training Strategy}&\multicolumn{1}{r@{\quad}}{Training Time}&
                   \multicolumn{1}{r@{\quad}}{mPrecision}&
                   \multicolumn{1}{r@{\quad}}{mRecall}&
                   \multicolumn{1}{r@{\quad}}{mF1}&
                   \multicolumn{1}{r}{mIoU}\\
\multicolumn{1}{l}{
                   }&\multicolumn{1}{|r@{\quad}}{(h)}&
                   \multicolumn{1}{r@{\quad}}{(\%)}&
                   \multicolumn{1}{r@{\quad}}{(\%)}&
                   \multicolumn{1}{r@{\quad}}{(\%)}&
                   \multicolumn{1}{r}{(\%)}\\[2pt]
\hline\rule{-1.5pt}{12pt}
Training from scratch  &    1.38  &  61.58  &  54.93  &  58.07  &  43.25 \\
\,Refining the rear-end  & $\boldsymbol{0.13}$ & 79.28  & 72.78  & 75.89  & 59.78\\
\,Refining the entire network  & 0.16  & $\boldsymbol{84.00}$  & $\boldsymbol{81.86}$  &  $\boldsymbol{82.92}$  & $\boldsymbol{70.94}$ \\[2pt]
\hline
\end{tabular}
\label{Tab:TransferLearning}
\end{table}


Training the network from scratch took 1.38 hours, and the network performance (61.58 \% mPrecision, 54.93\% mRecall, 58.07\% mF1-score, and 43.25\% mIoU) is well below satisfactory. Refining only the rear-end of the pre-trained network took only 0.13 hours to transfer the capability of an existing HRNetV2 to the new task with bridge elements. The transferred HRNetV2 reached a better performance level (79.28\% mPrecision, 72.78\% mRecall, and 75.89\% mF1-score, and 59.78\% mIoU). Compared to refining only the rear-end of the network, refining the entire pre-trained network took only two more minutes but markedly improved the performance (4.75 points gain on mPrecision, 9.08 points on mRecall, 7.03 points on mF1, and 11.16 points on mIoU).

\subsection{Comparison of State-of-the-Art Networks}
Five state-of-the-art image segmentation methods, U-Net \cite{unet}, Pyramid Scene Parsing Network (PSPNet) \cite{pspnet}, DeepLabv3+ \cite{deeplabv3+}, Mask Region-based Convolutional Neural Network (Mask R-CNN) \cite{mask}, and HRNet, were implemented to compare their performance in detecting and segmenting bridge elements from inspection images. The training sample size for this experiment is 70. Results from the comparison are summarized in Table \ref{Tab:models}. With similar training time to U-Net and PSPNet, HRNetV2 with the backbone W32 achieves the best results: 84.00\% mPrecision, 81.86\% mRecall, 82.92\% mF1, and 70.94\% mIoU. The gain of HRNetV2-W32 over U-Net (ResNet-50) on mIoU is 6.12 points and 4.16 points over PSPNet (ResNet-50). Compared with DeepLabv3+ with Xception, HRNetV2-W32 reduces the training time by about 80\% (0.58 h), and it achieves a much better performance: 5.15 points gain on mPrecision, 8.38 on mRecall, 6.85 points on mF1, and 10.8 points on mIoU. Mask R-CNN did not perform well on this task, which will be illustrated by examples and discussed later.


\begin{table}[htbp]
\caption{Comparison of the State-of-the-Art Networks for Segmentation}
\begin{tabular}{l@{\quad}l@{\quad}|r@{\quad}r@{\quad}r@{\quad}r@{\quad}rl}
\hline\rule{-2pt}{12pt}
\multirow{2}{*}{Network}&\multirow{2}{*}{Backbone}&
                   \multicolumn{1}{r@{\quad}}{Training}&
                   \multicolumn{1}{r@{\quad}}{mPrecision}&
                   \multicolumn{1}{r@{\quad}}{mRecall}&
                   \multicolumn{1}{r@{\quad}}{mF1}&
                   \multicolumn{1}{r}{mIoU}\\
\multicolumn{1}{l}{}&\multicolumn{1}{l}{}&
                   \multicolumn{1}{|r@{\quad}}{time (h)}&
                   \multicolumn{1}{r@{\quad}}{(\%)}&
                   \multicolumn{1}{r@{\quad}}{(\%)}&
                   \multicolumn{1}{r@{\quad}}{(\%)}&
                   \multicolumn{1}{r}{(\%)}\\[2pt]
\hline\rule{-1.5pt}{12pt}
U-Net & VGG16 &   $\boldsymbol{0.06}$  &  74.34 &  65.53  &  69.66  &  54.71 \\
\,U-Net & ResNet-50  & 0.16  & 83.15  & 75.40  &  79.09  & 64.82 \\
\,PSPNet & MobileNetV2  & 0.19 & 74.12  & 68.22  & 71.05  & 56.77\\
\,PSPNet & ResNet-50  & 0.10 & 82.82  & 77.75  & 80.20  & 66.78\\
\,DeepLabv3+ & MobileNetV2  & 0.29 & 80.01 &  73.41  & 76.57  & 61.21\\
\,DeepLabv3+ & Xception  & 0.74 & 78.85  & 73.48  & 76.07  & 60.14\\
\,Mask R-CNN & ResNet-50  & 0.43 & 46.97  & 37.00  & 41.39  & 27.56\\
\,Mask R-CNN & ResNet-101  & 0.48 & 47.83  & 40.39  & 43.80  & 28.04\\
\,HRNetV2 & HRNetV2-W18  & 0.16 & 82.91  & 74.86  & 78.68  & 64.52\\
\,HRNetV2 & HRNetV2-W32  & 0.16 & $\boldsymbol{84.00}$ & $\boldsymbol{81.86} $& $\boldsymbol{82.92}$ & $\boldsymbol{70.94}$\\
\,HRNetV2 & HRNetV2-W48  & 0.17 & 83.43  & 78.67  & 80.93  & 68.37\\[2pt]
\hline
\end{tabular}
\label{Tab:models}
\end{table}

\subsection{Impact of Training Sample Size}

Although HRNetV2-W32 performed the best among the state-of-the-art networks, the performance may be further improved if more training data becomes available. To examine the impact of training sample size and to refine the performance, this study refined the pre-trained HRNetV2-W32 using three training datasets in different sizes, summarized in Table \ref{Tab:Samplesize}.
\begin{table}[htbp]
\caption{Impact of the Training Sample Size on HRNetV2-W32}
\centering
\begin{tabular}{r|@{\quad}r@{\quad}r@{\quad}r@{\quad}r@{\quad}rl}
\hline\rule{-2pt}{12pt}
\multirow{2}{*}{Sample size}&\multicolumn{1}{|l@{\quad}}{Training time}&
                   \multicolumn{1}{r@{\quad}}{mPrecision}&
                   \multicolumn{1}{r@{\quad}}{mRecall}&
                   \multicolumn{1}{r@{\quad}}{mF1}&
                   \multicolumn{1}{r}{mIoU}\\
\multicolumn{1}{l|@{\quad}}{}&\multicolumn{1}{r@{\quad}}{(h)}&
                   \multicolumn{1}{r@{\quad}}{(\%)}&
                   \multicolumn{1}{r@{\quad}}{(\%)}&
                   \multicolumn{1}{r@{\quad}}{(\%)}&
                   \multicolumn{1}{r}{(\%)}\\[2pt]
\hline\rule{0pt}{12pt}
70 & $\boldsymbol{0.16}$  & 84.00  & 81.86  &  82.92  & 70.94  \\
100  & 0.32  & 91.03  & 91.29  &  91.16  & 83.85 \\
130   & 0.76 & $\boldsymbol{92.49}$  & $\boldsymbol{92.85}$  & $\boldsymbol{92.67}$  & $\boldsymbol{86.33}$\\[2pt]
\hline
\end{tabular}
\label{Tab:Samplesize}
\end{table}



\subsection{Class-wise Performance}

To further assess the performance of HRNetV2-32 across different classes of bridge elements, the class-level results achieved by the training dataset of 130 images are shown in Table \ref{Tab:classperform}. The table shows the network has an unequal ability to segment different classes of structural elements. For the background class, its IoU (79.34\%) and Recall (84.18\%) are the lowest among all classes, but its Precision is relatively high (93.24\%). The result indicates the background segmentation has more false negatives than false positives, meaning that the model has cautious but accurate background predictions. The possible cause could be image annotators identifying the dark areas in the inspection images as background while the model still could segment it even under poor lighting conditions. The Precision values of bearing, bracing, and substructure (87.05$\sim$88.25\%) are the lowest among all elements. Their Recall and IoU values are higher than the background class but still not the best. It means the model has loose predictions for these element classes, resulting in a high false-positive result. The high false-positive rate might be because these elements have irregular shapes compared with others. They mix with the cluttered and dynamic background in inspection images, especially for bracing that has a cross shape. The evaluation metrics of the deck, floor beam, and girder are the highest, which means the model performs well on these elements due to their relatively regular shapes.
\begin{table}[htbp]
\caption{Class-level Performance of HRNetV2-W32}
\centering
\begin{tabular}{l|@{\quad}r@{\quad}r@{\quad}r@{\quad}r@{\quad}rl}
\hline\rule{-2pt}{12pt}
\multirow{1}{*}{Element}&\multicolumn{1}{r@{\quad}}{Precision}&\multicolumn{1}{r@{\quad}}{Recall}&\multicolumn{1}{r@{\quad}}{F1}&\multicolumn{1}{r@{\quad}}{IoU}\\
&(\%)&(\%)&(\%)&(\%)\\
\hline\rule{-1.5pt}{12pt}
Background & 93.24 & 84.18 & 88.48 & 79.34 \\
\,Bearing & 88.25 & 92.62 & 90.38 & 82.45 \\
\,Bracing & 87.05 & 92.02 & 89.47 & 80.94 \\
\,Deck & $\boldsymbol{98.43}$ & 91.93 & 95.07 & 90.60 \\
\,Floor beam & 95.47 & 94.29 & 94.88 & 90.24 \\
\,Girder & 97.80 & $\boldsymbol{97.95}$ & $\boldsymbol{97.87}$ & $\boldsymbol{95.84}$ \\
\,Substructure & 87.20 & 96.98 & 91.83 & 84.90 \\[2pt]
\hline
\end{tabular}
\label{Tab:classperform}
\end{table}

\subsection{Effects of Data Augmentation and Class Weights}

Class imbalance could be a reason for the network's unequal performance across classes. This problem has a more noticeable effect on a small-size training dataset. This study employed data augmentation techniques to increase the diversity and representation of the training dataset, thus improving the network's generalization ability. These include the horizontal flip, rotation ($\pm10^{\circ}$), scale transformation, and random image intensity noise. Modifying the loss function using class-level weights is another possible solution. Weights for the classes are calculated as the inverse proportion of respective pixel or instance numbers. Table \ref{Tab:aug} shows the effectiveness of data augmentation and class weights. Data augmentation was particularly helpful for boosting the network's performance. Adding class weights onto the loss function in addition to the data augmentation can slightly improve some performance metrics, but it is marginal on this small dataset.


\begin{table}[htbp]
\caption{Effects of data augmentation and class weights}
\centering
\begin{tabular}{l@{\quad}l|r@{\quad}r@{\quad}r@{\quad}r@{\quad}rl}
\hline
\multicolumn{1}{l@{\quad}}{Data}&\multicolumn{1}{l|}{Class}&\multicolumn{1}{r@{\quad}}{Training time}&
\multicolumn{1}{r@{\quad}}{mPrecision}&
\multicolumn{1}{r@{\quad}}{mRecall}&
\multicolumn{1}{r@{\quad}}{mF1}&
\multicolumn{1}{r}{mIoU}\\
\multicolumn{1}{l@{\quad}}{augmentation
}&\multicolumn{1}{l|}{weights}&\multicolumn{1}{r@{\quad}}{(h)}&
\multicolumn{1}{r@{\quad}}{(\%)}&
\multicolumn{1}{r@{\quad}}{(\%)}&
\multicolumn{1}{r@{\quad}}{(\%)}&
\multicolumn{1}{r}{(\%)}\\[2pt]
\hline\rule{-1.5pt}{12pt}
no & no & 0.19  &  26.33  & 33.30  & 29.41  & 19.16  \\
\,yes & no  & $\boldsymbol{0.16}$  & $\boldsymbol{84.00}$  & 81.86  &  82.92  & 70.94 \\
\,yes &yes  & 0.19 &  83.22  & $\boldsymbol{83.36}$  & $\boldsymbol{83.29}$  & $\boldsymbol{71.40}$\\[2pt]
\hline
\end{tabular}
\label{Tab:aug}
\end{table}

%

\subsection{Qualitative examples}
%

\begin{figure}[htb]
\centering
\includegraphics[width=10.5cm]{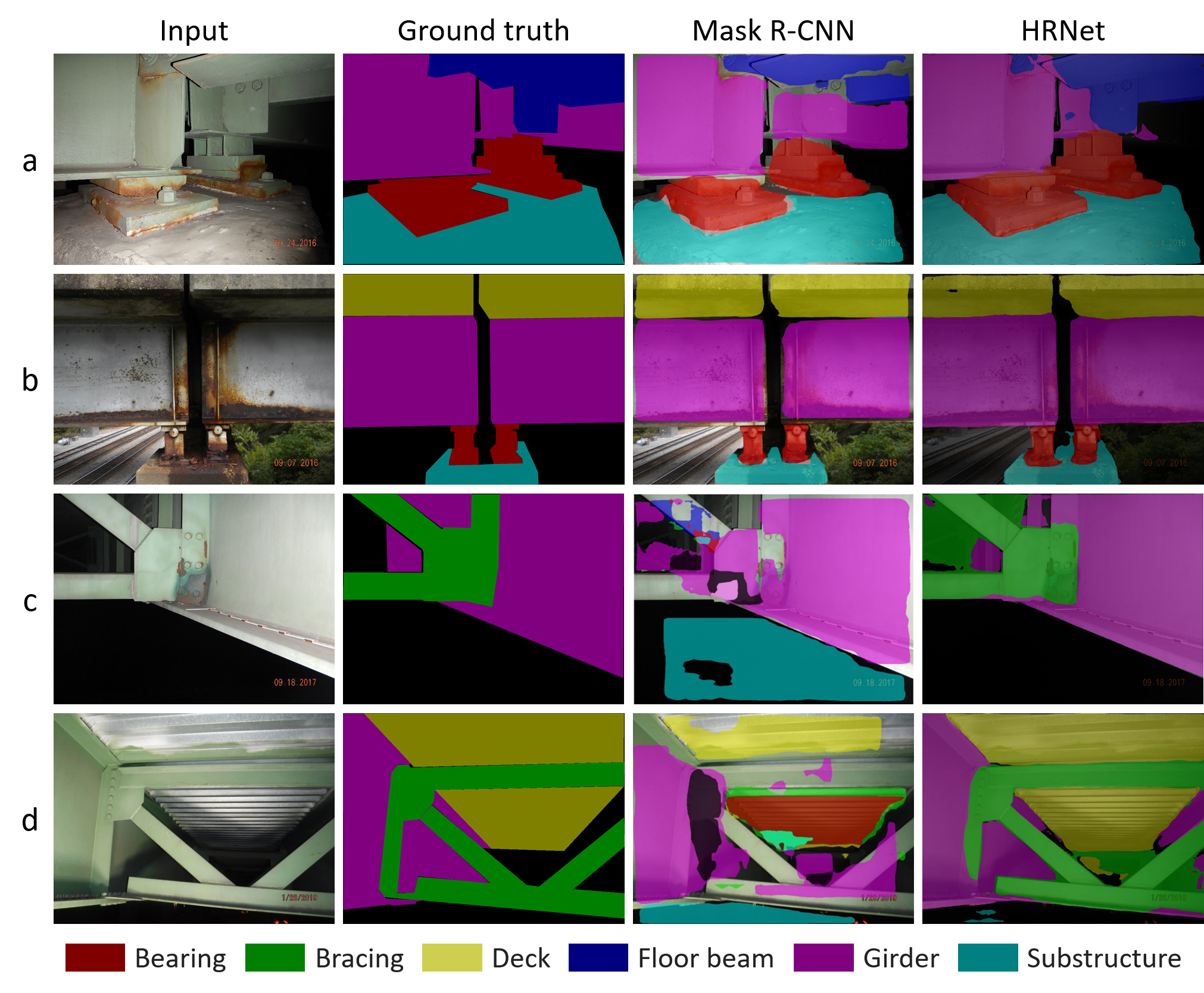}
\caption{Examples}
\label{Fig_Examples}
\end{figure}

Fig. \ref{Fig_Examples} illustrates four examples wherein the bridge elements occupy most areas of the images. HRNetV2-W32 performed well on all examples by maintaining relatively high resolution in segmenting bridge elements. However, some false negatives are present in example a and some false positives in c and d. Mask R-CNN provided reasonable predictions in examples a and b, wherein bridge elements have relatively regular shapes. However, it fails to segment large elements and those of irregular shapes in examples c and d, probably because some anchor boxes are almost as big as the image and are close to each other.


%
\section{Conclusions}
This study demonstrated that HRNet-W32 is a suitable DNN for segmenting multiclass structural elements from bridge inspection images, especially when elements are large or have complex shapes in images. With data augmentation and a training dataset of 130 images, a pre-trained HRNet has been transferred to this image analysis task and achieved a promising result. An immediate extension is to develop a multi-tasking deep learning framework to further quantify surface defects based on the segmented structural elements.


\section*{Acknowledgement}
Qin and Zhang receive financial support from the National Science Foundation Award ECCS-2026357. Yin and Karim receive financial support from the National Science Foundation Award ECCS-2025929.
%
%

\end{document}